\documentclass{article}

\usepackage[nonatbib, final]{nips_2016}
\usepackage[utf8]{inputenc} 
\usepackage[T1]{fontenc}    
\usepackage{hyperref}       
\usepackage{url}            
\usepackage{booktabs}       
\usepackage{amsfonts}       
\usepackage{nicefrac}       
\usepackage{microtype}      
\usepackage{graphicx}
\usepackage{caption}
\usepackage{subcaption}
\usepackage{color}
\usepackage{amsmath}
\title{Deep Reinforcement Learning for Autonomous Driving}

\author{
	Sen Wang\\
	Carnegie Mellon University\\
	5000 Forbes Ave, \\
	Pittsburgh, PA 15213\\
    \texttt{senw@andrew.cmu.com} 
	\And
	Daoyuan Jia \\
	Carnegie Mellon University\\
	5000 Forbes Ave, \\
    Pittsburgh, PA 15213\\
  \texttt{daoyuanj@andrew.cmu.edu} 
  	\And
	Xinshuo Weng \\
	Carnegie Mellon University\\
	5000 Forbes Ave, \\
    Pittsburgh, PA 15213\\
  \texttt{xinshuow@andrew.cmu.edu} \\
}

\begin{document}
\maketitle

\begin{abstract}
Reinforcement learning has steadily improved and outperform human in lots of traditional games since the resurgence of deep neural network. However, these success is not easy to be copied to autonomous driving because the state spaces in real world are extreme complex and action spaces are continuous and fine control is required. Moreover, the autonomous driving vehicles must also keep functional safety under the complex environments. To deal with these challenges, we first adopt the deep deterministic policy gradient (DDPG) algorithm, which has the capacity to handle complex state and action spaces in continuous domain. We then choose The Open Racing Car Simulator (TORCS) as our environment to avoid physical damage. Meanwhile, we select a set of appropriate sensor information from TORCS and design our own rewarder. In order to fit DDPG algorithm to TORCS, we design our network architecture for both actor and critic inside DDPG paradigm. To demonstrate the effectiveness of our model, We evaluate on different modes in TORCS and show both quantitative and qualitative results.
\end{abstract}

\section{Introduction}

Autonomous driving \cite{Memon2016} is an active research area in computer vision and control systems. Even in industry, many companies, such as Google, Tesla, NVIDIA \cite{bojarski2016end}, Uber and Baidu, are also devoted to developing advanced autonomous driving car because it can really benefit human's life in real world. On the other hand, deep reinforcement learning technique has been successfully applied with great success to a variety of games \cite{mnih2013playing} \cite{mnih2015human}. The success of deep reinforcement learning algorithm proves that the control problems in real-world environment could be naturally solved by optimizing policy-guided agents in high-dimensional state and action space. In particular, state spaces are often represented by image features obtained from raw images in vision control systems.

However, the current success achieved by deep reinforcement learning algorithms mostly happens in scenarios where controller has only discrete and limited action spaces and there is no complex content in state spaces of the environment, which is not the case when applying deep reinforcement learning algorithms to autonomous driving system. For example, there are only four actions in some Atari games such as SpaceInvaders and Enduro. For game Go, the rules and state of boards are very easy to understand visually even though spate spaces are high-dimensional. In such cases, vision problems are extremely easy to solve, then the agents only need to focus on optimizing the policy with limited action spaces. But for autonomous driving, the state spaces and input images from the environments contain highly complex background and objects inside such as human which can vary dynamically and behave unpredictably. These involve in lots of difficult vision tasks such as object detection, scene understanding, depth estimation. More importantly, our controller has to act correctly and fast in such difficult scenarios to avoid hitting objects and keep safe.

A straightforward way of achieving autonomous driving is to capture the environment information by using precise and robust hardwares and sensors such as Lidar and Inertial Measurement Unit (IMU). These hardware systems can reconstruct the 3D information precisely and then help vehicle achieve intelligent navigation without collision using reinforcement learning. However, there hardwares are very expensive and heavy to deploy. More importantly, they only tell us the 3D physical surface of the world instead of understanding the environment, which is not really intelligent. Both these reasons from hardware systems limit the popularity of autonomous driving technique. 

One alternative solution is to combine vision and reinforcement learning algorithm and then solve the perception and navigation problems jointly. However, the perception problem is very difficult to solve because our world is extreme complex and unpredictable. In other words, there are huge variance in the world, such as color, shape of objects, type of objects, background and viewpoint. Even stationary environment is hard to understand, let alone the environment is changing as the autonomous vehicle is running. Meanwhile, the control problem is also challenging in real world because the action spaces is continuous and different action can be executed at the same time. For example, for smoother turning, We can steer and brake at the same time and adjust the degree of steering as we turn. More importantly, A safe autonomous vehicle must ensure functional safety and be able to deal with urgent events. For example, vehicles need to be very careful about crossroads and unseen corners such that they can act or brake immediately when there are children suddenly running across the road.

In order to achieve autonomous driving, people are trying to leverage information from both sensors and vision algorithms. 
Lots of synthetic driving simulators are made for learning the navigation policy without physical damage. Meanwhile, people are developing more robust and efficient reinforcement learning algorithm \cite{silver2014deterministic, mnih2016asynchronous, wang2015dueling, van2016deep, Abbeel2016} in order to successfully deal with situations with real-world complexity. In this project, we are trying to explore and analyze the possibility of achieving autonomous driving within synthetic simulators. 

In particular, we adopt deep deterministic policy gradient (DDPG) algorithm \cite{lillicrap2015continuous}, which combines the ideas of deterministic policy gradient, actor-critic algorithms and deep Q-learning. We choose The Open Racing Car Simulator (TORCS) as our environment to train our agent. In order to learn the policy in TORCS, We first select a set of appropriate sensor information as inputs from TORCS. Based on these inputs, we then design our own rewarder inside TORCS to encourage our agent to run fast without hitting other cars and also stick to the center of the road. Meanwhile, in order to fit in TORCS environment, we design our own network architecture for both actor and critic used in DDPG algorithm. To demonstrate the effectiveness of our method, we evaluate our agent in different modes in TORCS, which contains different visual information.

\section{Related Work}

\textbf{Autonomous Driving.} Attempts for solving autonomous driving can track back to traditional control technique before deep learning era. Here we only discuss recent advances in autonomous driving by using reinforcement learning or deep learning techniques. Karavolos \cite{Karavolos2013} apply the vanilla Q-learning algorithm to simulator TORCS and evaluate the effectiveness of using heuristic during the exploration. Huval2015 \textit{et al.} \cite{Huval2015} propose a CNN-based method to decompose autonomous driving problem into car detection, lane detection task and evaluate their method in a real-world highway dataset. On the other hand, Bojarski \textit{et al.} \cite{bojarski2016end} achieve autonomous driving by proposing an end to end model architecture and test it on both simulators and real-world environments. Sharifzadeh2016 \textit{et al.} \cite{Sharifzadeh2016} achieve collision-free motion and human-like lane change behavior by using an inverse reinforcement learning approach. Different from prior works, Shalev-shwartz \textit{et al.} \cite{shwartz2016} model autonomous driving as a multi-agent control problem and demonstrate the effectiveness of a deep policy gradient method on a very simple traffic simulator. Seff and Xiao \cite{Seff2016} propose to leverage information from Google Map and match it with Google Street View images to achieve scene understanding prior to navigation. Recent works \cite{Yang2017, Chae2017, Isele2017} are mainly focus on deep reinforcement learning paradigm to achieve autonomous driving. In order to achieve autonomous driving in th wild, You \textit{et al.} \cite{You2017} propose to achieve virtual to real image translation and then learn the control policy on realistic images.

\textbf{Reinforcement Learning.} Existing reinforcement learning algorithms mainly compose of value-based and policy-based methods. Vanilla Q-learning is first proposed in \cite{watkins1989learning} and then become one of popular value-based methods. Recently lots of variants of Q-learning algorithm, such as DQN \cite{mnih2015human}, Double DQN \cite{van2016deep} and Dueling DQN \cite{wang2015dueling}, have been successfully applied to a variety of games and outperform human since the resurgence of deep neural networks. By leveraging the advantage functions and ideas from actor-critic methods \cite{Konda1999}, A3C \cite{mnih2016asynchronous} further improve the performance of value-based reinforcement learning methods.

Different from value-based methods, policy-based methods learn the policy directly. In other words, policy-based methods output actions given current state. Silver \textit{et al.} \cite{silver2014deterministic} propose a deterministic policy gradient algorithm to handle continuous action spaces efficiently without losing adequate exploration. By combining idea from DQN and actor-critic, Lillicrap \textit{et al.} \cite{lillicrap2015continuous} then propose a deep deterministic policy gradient method and achieve end-to-end policy learning. Very recently, PGQL \cite{ODonoghue2017} is proposed and can even outperform A3C by combining off-policy Q-learning with policy gradient. More importantly, in terms of autonomous driving, action spaces are continuous and fine control is required. All these policy-gradient methods can naturally handle the continuous action spaces. However, adapting value-based methods, such as DQN, to continuous domain by discretizing continuous action spaces might cause curse of dimensionality and can not meet the requirements of fine control. 

\section{Methods}
In autonomous driving, action spaces are continuous. For example, steering can vary from $-90^{\circ}$ to $90^{\circ}$ and acceleration can vary from 0 to 300km. This continuous action space will lead to poor performance for value-based methods. We thus use policy-based methods in this project. Meanwhile, random exploration in autonomous driving might lead to unexpected performance and terrible consequence. So we determine to use Deep Deterministic Policy Gradient (DDPG) algorithm, which uses a deterministic instead of stochastic action function. In particular, DDPG combines the advantages of deterministic policy gradient algorithm, actor-critics and deep Q-network.

In this section, we describe deterministic policy gradient algorithm and then explain how DDPG combines it with actor-critic and ideas from DQN together. Finally we explain how we fit our model in TORCS and design our reward signal to achieve autonomous driving in TORCS.

\subsection{Deterministic Policy Gradient (DPG)}

A stochastic policy can be defined as:
\begin{equation}
  \pi_{\theta}=P[a|s;\theta]
\end{equation}

Then the corresponding gradient is:
\begin{equation}
  \nabla_{\theta}J(\pi_{\theta})=E_{s\sim p^{\pi},a\sim \pi_{\theta}}[\nabla_{\theta}log\pi_{\theta}(a|s)Q^{\pi}(s,a)]
\end{equation}

This shows that the gradient is an expectation of possible states and actions. Thus in principle, in order to obtain an approximate estimation of the gradient, we need to take lots of samples from the action spaces and state spaces. Fortunately, mapping is fixed from state spaces to action spaces in deterministic policy gradient, so we do not need to integrate over whole action spaces. Thus deterministic policy gradient algorithm needs much fewer data samples to converge over stochastic policy gradient. Deterministic policy gradient is the expected gradient of the action-value function, so it can be estimated much efficiently than stochastic version.

\begin{figure}[!t]
  \centering
  		\includegraphics[width=0.5\textwidth]{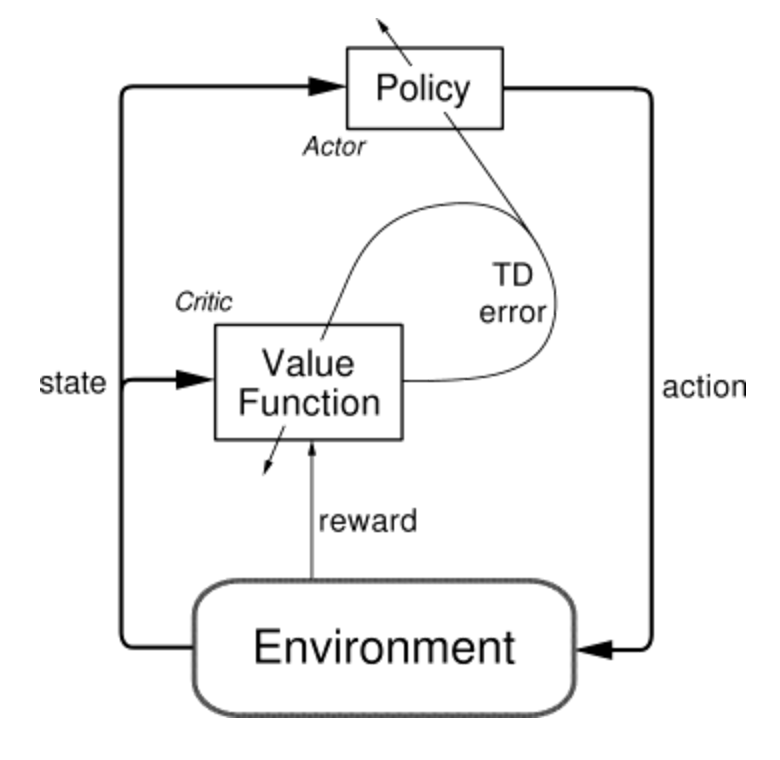}
   		\caption{Overall work flow of actor-critic paradigm.}
  \label{fig:ac}
\end{figure}

In order to explore the environment, DPG algorithm achieves off-policy learning by borrowing ideas from actor-critic algorithms. An overall work flow of actor-critic algorithms is shown in Figure \ref{fig:ac}. In particular, DPG composes of an actor, which is the policy to learn, and a critic, which estimating Q value function. Essentially, the actor produces the action a given the current state of the environment s, while the critic produces a signal to criticizes the actions made by the actor. Then the critic is updated by TD learning and the actor is updated by policy gradient. Assume the function parameter for critic is $w$ and the function parameter for Actor is $\theta$, the gradient for deterministic policy is:
\begin{equation}
	\nabla_{\theta}J(\mu_{\theta})=E_{s\sim p^{\mu}}[\nabla_{\theta}\mu_{\theta}(s)\nabla_{a}Q^{\mu}(s,a)|a=\mu_{\theta}(s)]
\end{equation}

For exploration stochastic policy $\beta$ and off deterministic policy $\mu_{\theta}(s)$, we can derive the off-policy policy gradient:
\begin{equation}
	\nabla_{\theta}J_{\beta}(\mu_{\theta})=E_{s\sim p^{\beta}}[\nabla_{\theta}\mu_{\theta}(s)\nabla_{a}Q^{\mu}(s,a)|a=\mu_{\theta}(s)]
\end{equation}

Notice that the formula does not have importance sampling factor. The reason is that the importance sampling is to approximate a complex probability distribution with a simple one. But the output of the policy here is a value instead of a distribution. The Q value in the formula corresponds to the critics and is updated by TD(0) learning. Given the policy gradient direction, we can derive the update process for Actor-Critic off-policy DPG:
\begin{align}
\delta_{t}&=r_{t}+\gamma Q^{w}(s_{t+1},\mu_{\theta}(s_{t+1}))-Q^{w}(s_{t},a_{t})\\
w_{t+1}&=w_{t}+\alpha_{w}\delta_{t}\nabla_{w}Q^{w}(s_{t},a_{t})\\
\theta_{t+1}&=\theta_{t}+\alpha_{\theta}\nabla_{\theta}\mu_{\theta}(s_{t})\nabla_{a}Q^{w}(s_{t},a_{t})|a=\mu_{\theta}(s)
\end{align}

\subsection{Deep Deterministic Policy Gradient (DDPG)}
DDPG algorithm mainly follow the DPG algorithm except the function approximation for both actor and critic are represented by deep neural networks. Instead of using raw images as inputs, 
Torcs supports various type of sensor input other than images as observation. Here, we chose to take all sensor input listed in Table~\ref{tbl:sensor_input}, make it a 29 dimension vector. The action of the model is a 3 dimension vector for \textbf{Acceleration} (where 0 means no gas, 1 means full gas), \textbf{Brake} (where 0 means no brake, 1 full brake) and \textbf{Steering} (where -1 means max right turn and +1 means max left turn) respectively. 

\begin{figure}[!t]
  \centering
  \begin{subfigure}[b]{0.45\textwidth}
   \centering
  		\includegraphics[width=\textwidth]{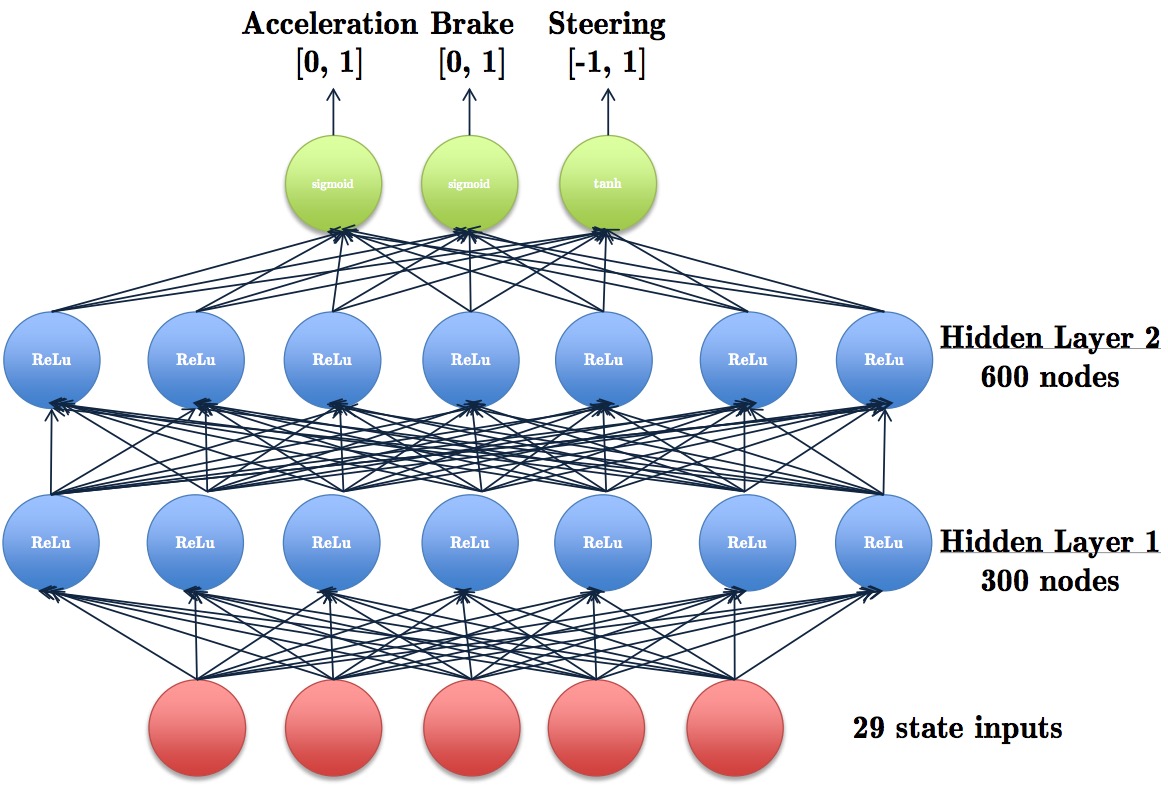}
   		\caption{Actor Network}
  \label{fig:ac_structure}
  \end{subfigure}
  \begin{subfigure}[b]{0.45\textwidth}
  \centering
  		\includegraphics[width=\textwidth]{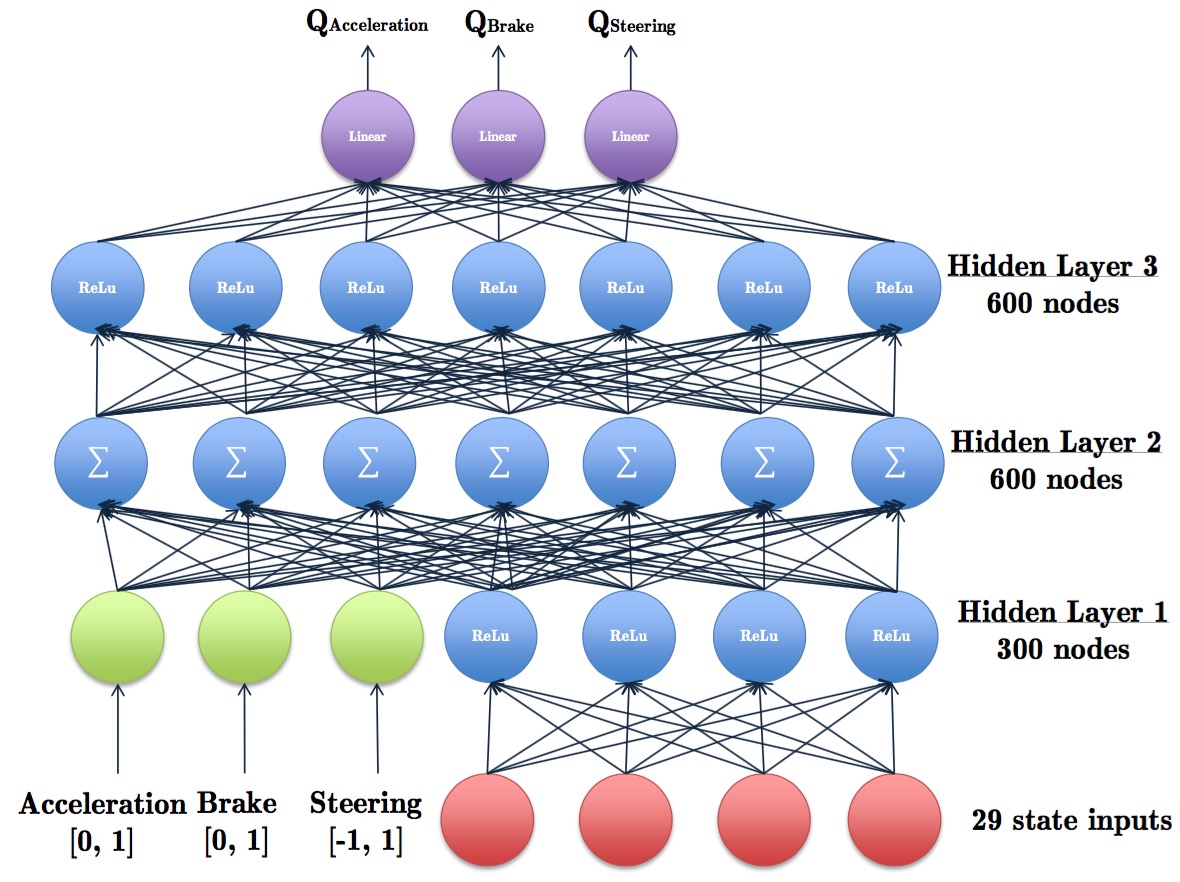}
   		\caption{Critic Network}
  \label{fig:c_structure}
  \end{subfigure}
  \caption{Actor and Critic network architecture in our DDPG algorithm.}
  \label{fig:actorcritic}
\end{figure}

The whole model is composed with an actor network and a critic network and is illustrated in Figure \ref{fig:actorcritic}. The actor network serves as the policy, and will output the action. Both hidden layers are comprised of ReLU activation function. The critic model serves as the Q-function, and will therefore take action and observation as input and output the estimation rewards for each of action. In the network, both previous action the actions are not made visible until the second hidden layer. The first and third hidden layers are ReLU activated, while the second merging layer computes a point-wise sum of a linear activation computed over the first hidden layer and a linear activation computed over the action inputs.

Meanwhile, in order to increase the stability of our agent, we adopt experience replay to break the dependency between data samples. A target network is used in DDPG algorithm, which means we create a copy for both actor and critic networks. Then these target networks are used for providing target values. The weights of these target networks are then updated in a fixed frequency. For actor and critic network, the parameter $w$ and $\theta$ are updated respectively by:
\begin{align}
\theta^{'}&=\tau \theta+(1-\tau)\theta^{'}\\
w^{'}&=\tau w+(1-\tau)w^{'}
\end{align}
\subsection{The Open Racing Car Simulator (TORCS)}
TORCH provides 18 different types of sensor inputs. After experiments we carefully select a subset of inputs, which is shown in Table \ref{tbl:sensor_input}.
\begin{table}[!t]
\centering
\begin{tabular}{|c|c|}
\hline
Name & Range(Unit)\\
\hline
ob.angle & [-pi,pi]\\
\hline
ob.track & (0,200)(meters)\\
\hline
ob.trackPos & $(-\inf,\inf)$ \\
\hline
ob.speedX & $(-\inf,\inf)$(km/h) \\
\hline
ob.speedY & $(-\inf,\inf)$(km/h) \\
\hline
ob.speedZ & $(-\inf,\inf)$(km/h) \\
\hline
\end{tabular}
\vspace{0.3cm}
\caption{Selected Sensor Inputs. }
\label{tbl:sensor_input}
\end{table}
\begin{itemize}
\item ob.angle is the angle between the car direction and the direction of the track axis. It reveals the car's direction to the track line. 
\item ob.track is the vector of 19 range finder sensors: each sensor returns the distance between the track edge and the car within a range of 200 meters. It let us know if the car is in danger of running into obstacle. 
\item ob.trackPos is the distance between the car and the track axis. The value is normalized w.r.t. to the track width: it is 0 when the car is on the axis, values greater than 1 or -1 means the car is outside of the track. We want the distance to the track axis to be 0. 
\item ob.speedX, ob.speedY, ob.speedZ is the speed of the car along the longitudinal axis of the car (good velocity), along the transverse axis of the car, and along the Z-axis of the car. We want the car speed along the axis to be high and speed vertical to the axis to be low.
\end{itemize}

\textbf{Reward Design} TORCS does not have internal rewarder, so we need to design our own reward function. The reward should not only encourage high speed along the track axis, but also punish speed vertical to the track axis as well as deviation from the track. We formulate our reward function as follows:
\begin{equation}
  R_{t} = V_{x}cos(\theta)-\alpha V_{x}sin(\theta)-\gamma|trackPos|-\beta V_{x}|trackPos|
\end{equation}

$V_{x}cos(\theta)$ denotes the speed along the track, which should be encouraged. $V_{x}sin(\theta)$ denotes the speed vertical to the track. |trackPos| measures the distance between the car and the track line. Both $|trackPos|$ and $V_{x}|trackPos|$ punish the agent when the agent deviates from center of the road. $\alpha,\beta,\gamma$ denote the weight for each reward term respectively.

\section{Experiments}
\subsection{Experimental Setting}	\label{sec:exp_setting}

We experiment the simple actor-critic algorithm on TORCS engine, and asynchronous actor-critic algorithm on OpenAI Universe. We choose TORCS as the environment for TORCS, since it was wrapped OpenAI-compatible interfaces. Other softwares include Anaconda 2.7, Keras and Tensorflow v0.12. All our experiments were made on an Ubuntu 16.04 machine, with 12 cores CPU, 64GB memory and 4 GTX-780 GPU (12GB Graphic memory in total).


We adapt the implementation and hyper-parameter choice from \cite{TorcsKeras}. Specifically, the replay buffer size is $100000$ state-action pairs, with a discount factor of $\gamma= 0.99$. The optimizer is Adam with learning rates of 0.0001 and 0.001 for the actor and critic respectively, and a batch-size of 32. Target networks are updated gradually with $\tau=0.001$.

\subsection{Experiment Analysis}
\begin{figure}[!t]
  \centering
  \begin{subfigure}[b]{0.45\textwidth}
    \centering
\includegraphics[width=\textwidth, height=3.8cm]{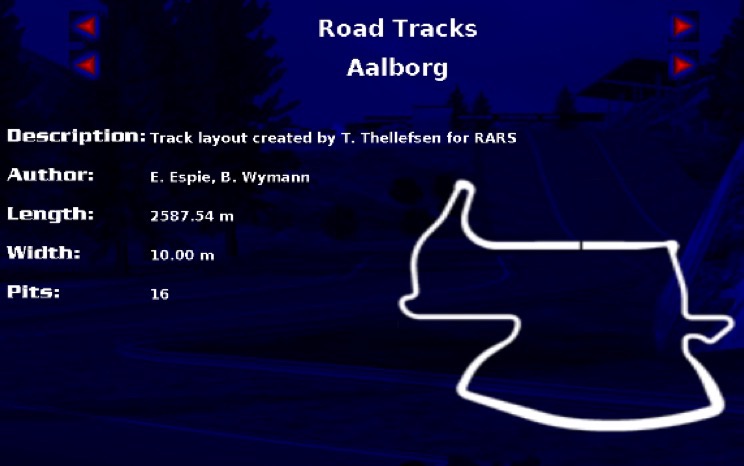}
   		\caption{Map Aalborg}
  \label{fig:aalborg}
  \end{subfigure}
  \begin{subfigure}[b]{0.45\textwidth}
    \centering
  		\includegraphics[width=\textwidth, height=3.8cm]{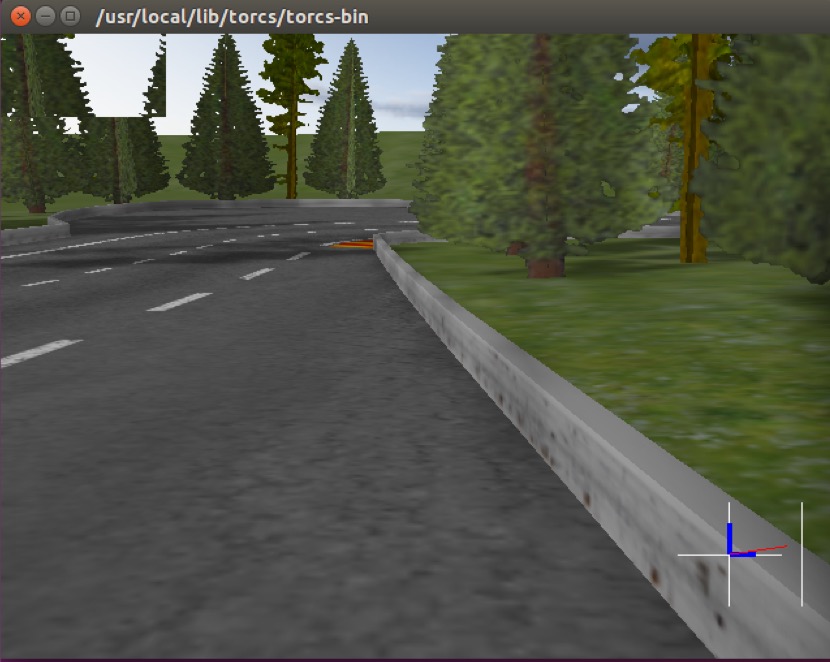}
   		\caption{Training Mode: shaky at beginning of training}
  \label{fig:a0}
  \end{subfigure}
    \begin{subfigure}[b]{0.45\textwidth}
    \centering
  		\includegraphics[width=\textwidth, height=4.3cm]{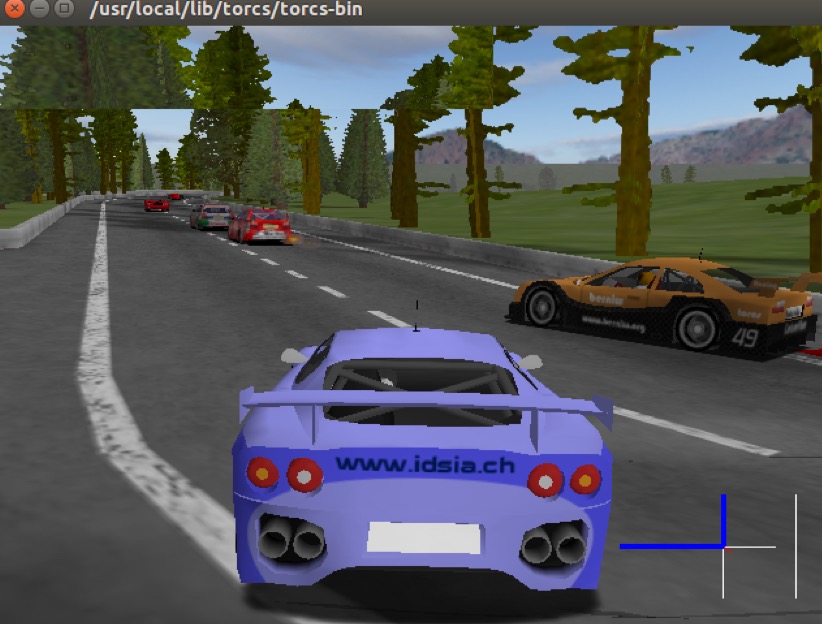}
   		\caption{Compete Mode: falling behind at beginning}
  \label{fig:a1}
  \end{subfigure}
  \begin{subfigure}[b]{0.45\textwidth}
    \centering
  		\includegraphics[width=\textwidth, height=4.3cm]{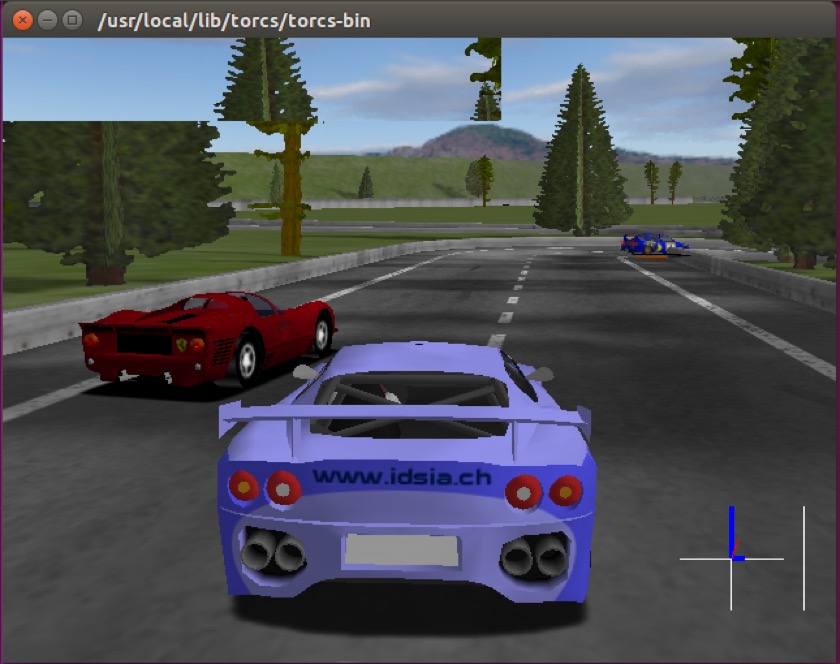}
   		\caption{Compete Mode: Overtake other cars at corner}
  \label{fig:a2}
  \end{subfigure}
  \caption{Train and evaluation on map Aalborg}
  \label{fig:aalborg_train_test}
\end{figure}

The TORCS engine contains many different modes. We can generally categorize them into two types: training mode and compete mode. In training mode, no other competitors in the view, and the view-angle is first-person as in Figure \ref{fig:a0}. In compete mode, we can add other computer-controlled AI into the game and racing with them, as shown in Figure \ref{fig:a1}. Notably, the existence of other competitors will affect the sensor input of our car. 


We train the game with about 200 episodes on map Aalborg in train mode, and evaluate the game in compete mode with 9 other competitors. Each episode terminates when the car rush out of the track or when the car orientated to the opposite direction. Therefore, the length of each episode is highly variated, and therefore a good model could make one episode infinitely. Thus, we also set the maximum length of one episode as 60000 iterations. The map is shown in Figure \ref{fig:aalborg}. In the train mode, the model is shaky at beginning, and bump into wall frequently (Figure \ref{fig:a0}), and gradually stabilize as training goes on. In evaluation (compete mode), we set our car ranking at 5 at beginning among all competitors. Therefore, our car fall behind 4 other cars at beginning (Figure \ref{fig:a1}). However, as the race continues, our car easily overtake other competitors in turns, shown in Figure \ref{fig:a2}. 

\begin{figure}[!t]
    \includegraphics[width=0.24\textwidth,height=8.5em]{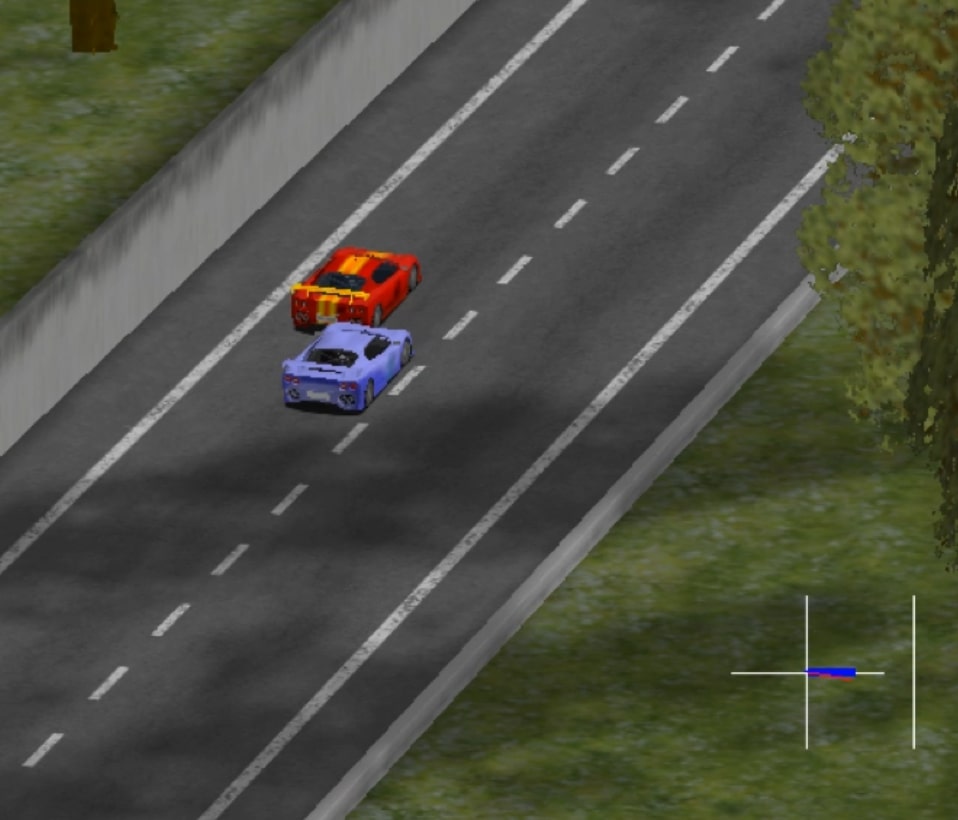}
    \includegraphics[width=0.24\textwidth,height=8.5em]{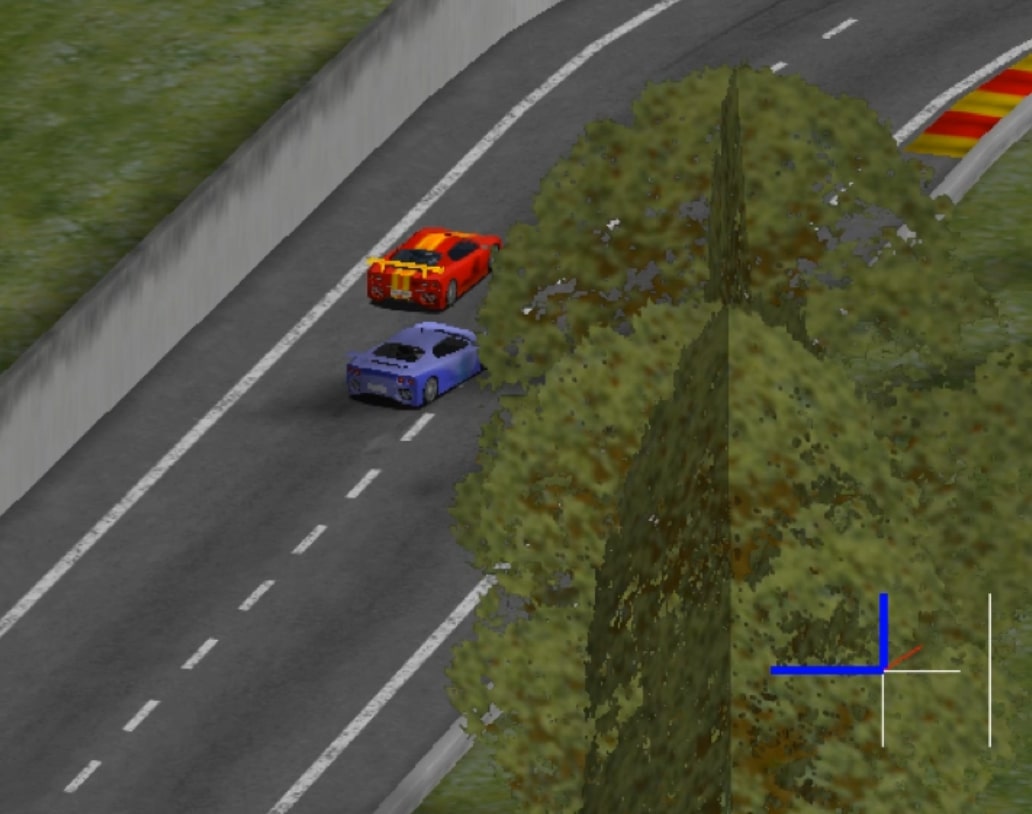}            
    \includegraphics[width=0.24\textwidth,height=8.5em]{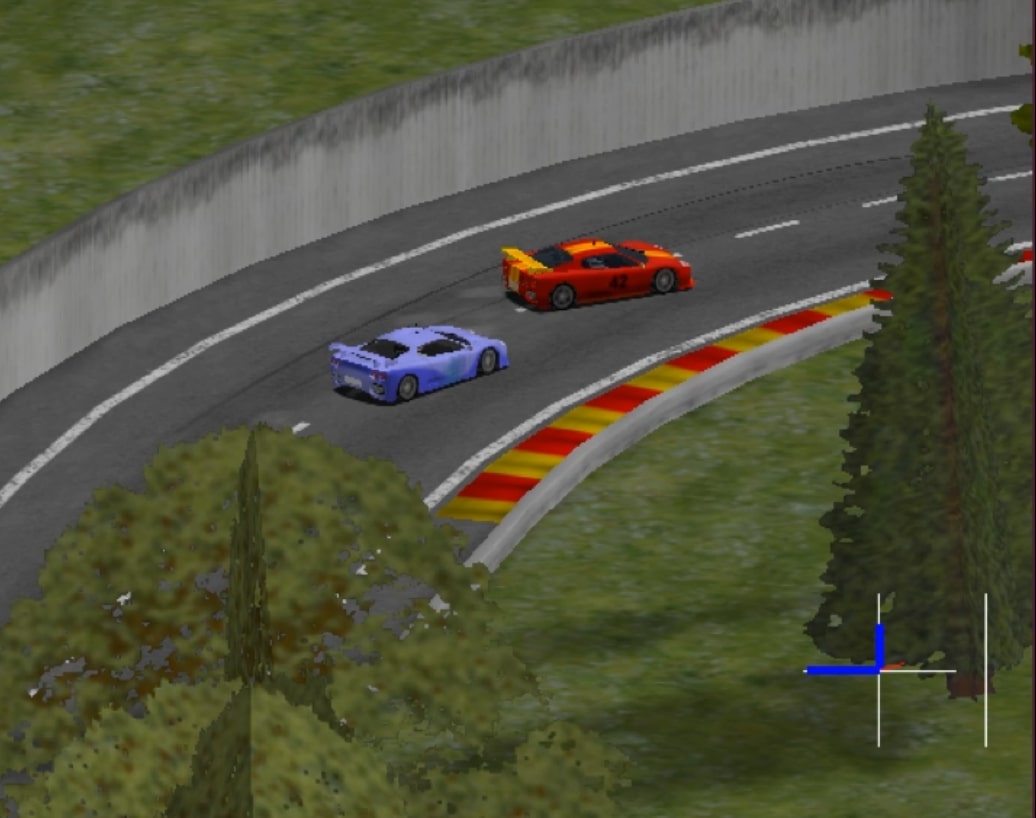}
    \includegraphics[width=0.24\textwidth,height=8.5em]{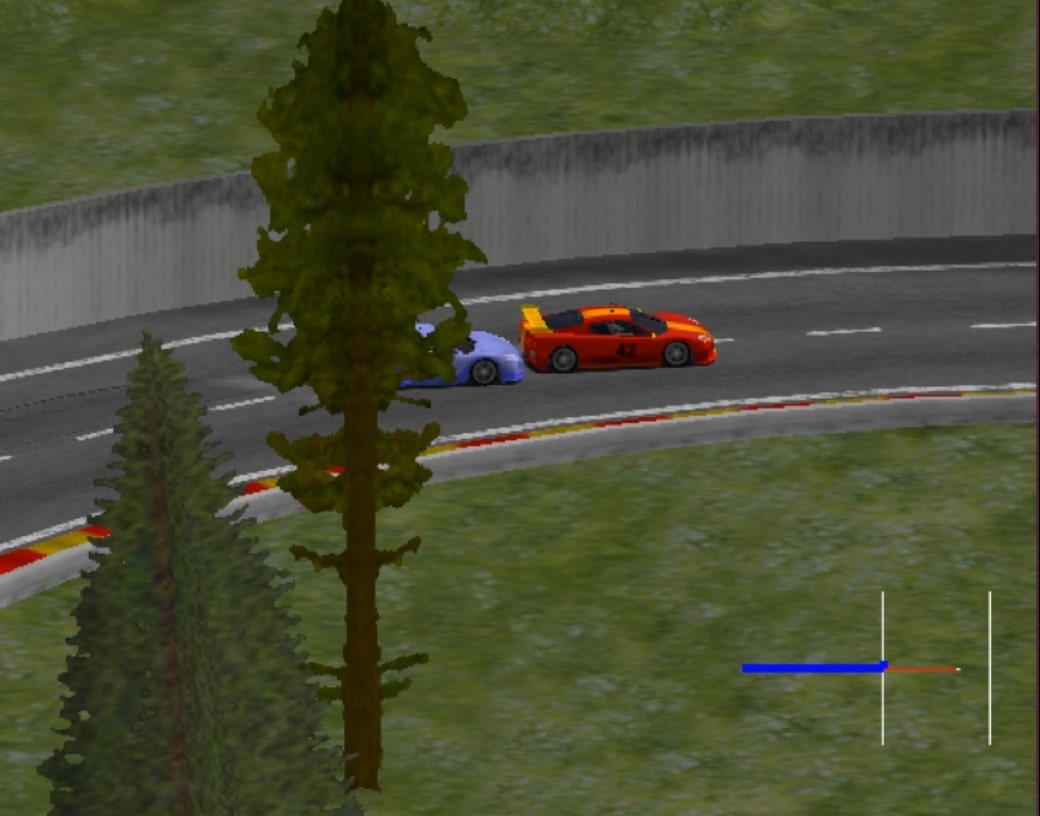}\\
    \includegraphics[width=0.24\textwidth,height=8.5em]{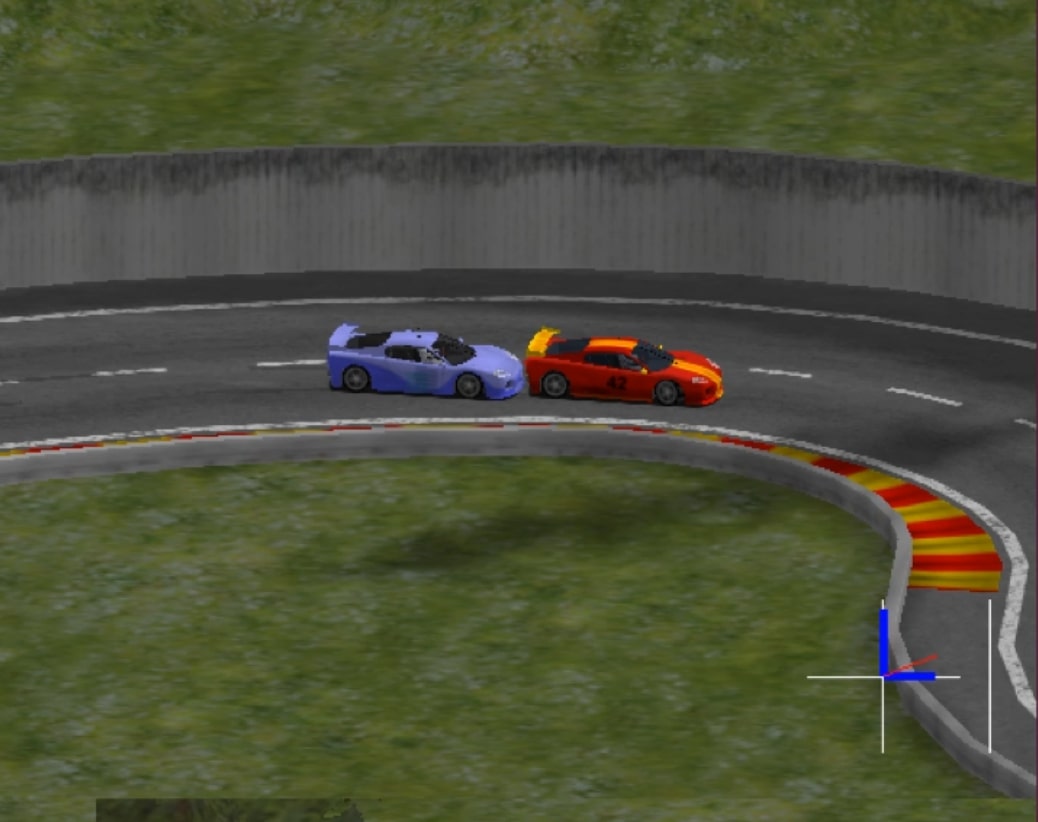}
    \includegraphics[width=0.24\textwidth,height=8.5em]{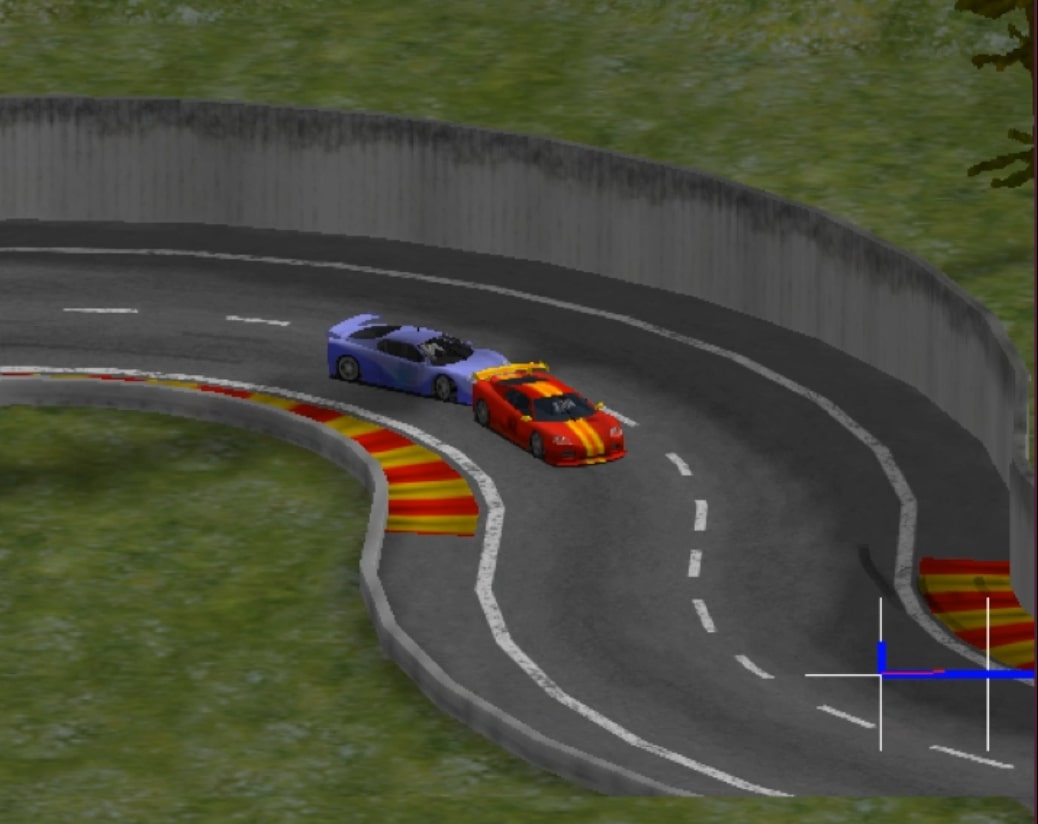}
    \includegraphics[width=0.24\textwidth,height=8.5em]{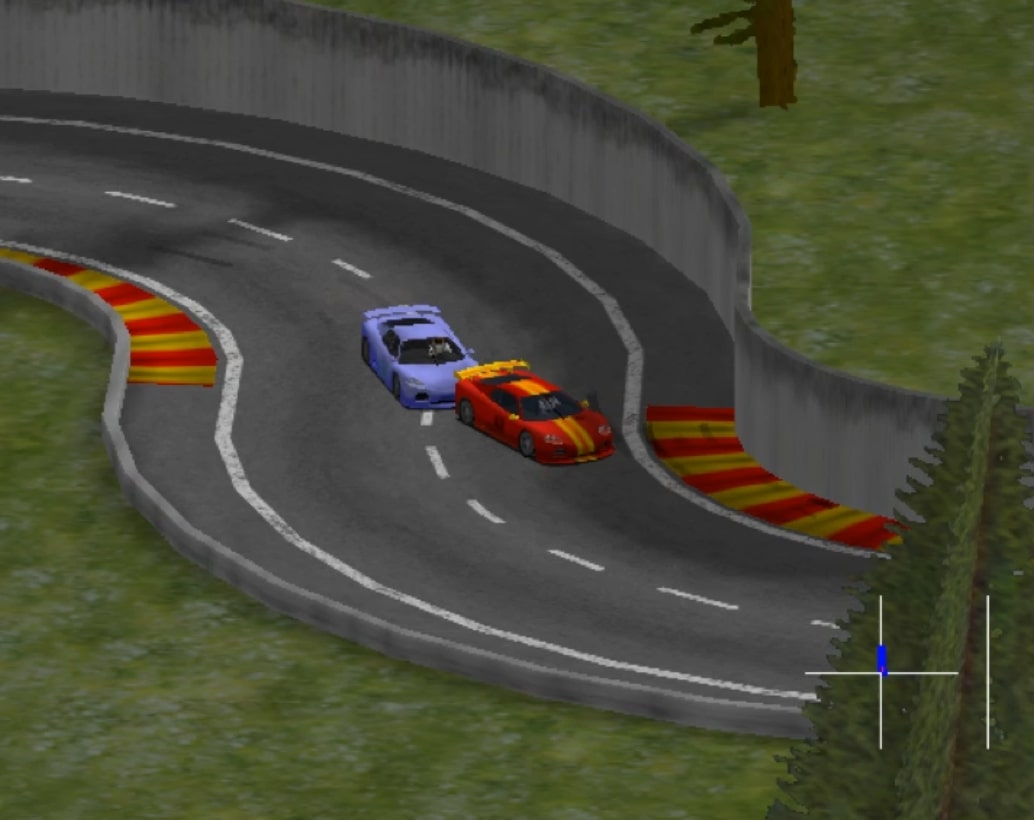}
    \includegraphics[width=0.24\textwidth,height=8.5em]{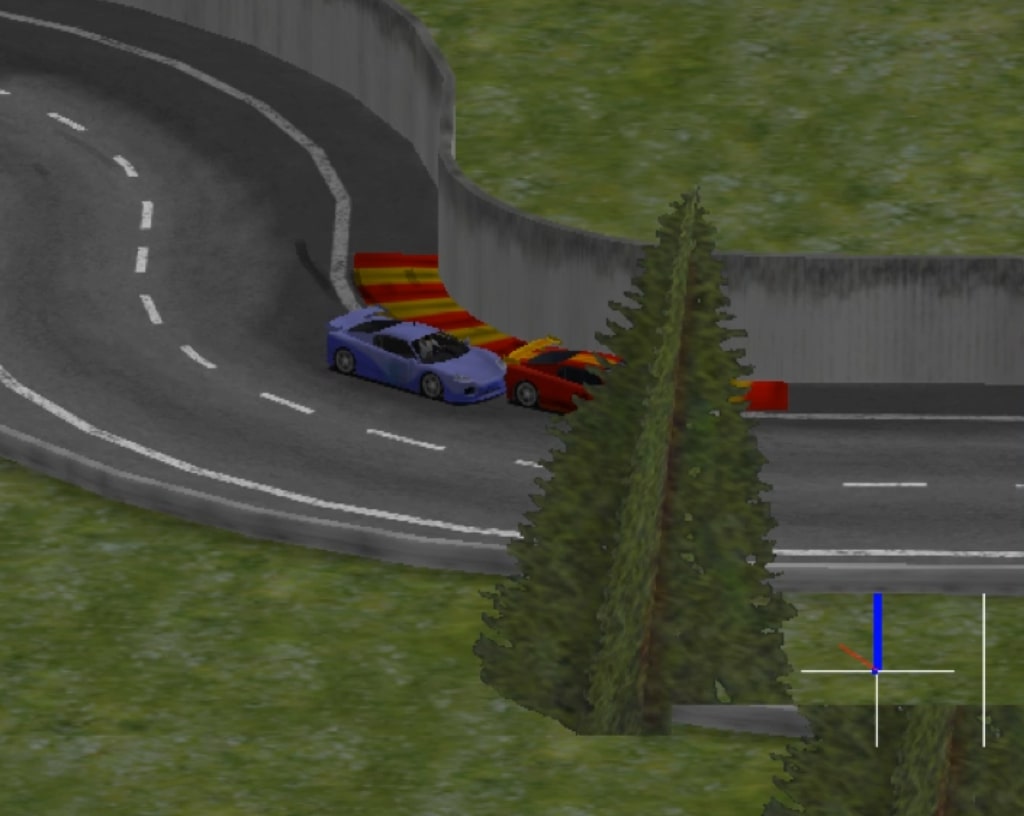}\\
    \includegraphics[width=0.24\textwidth,height=8.5em]{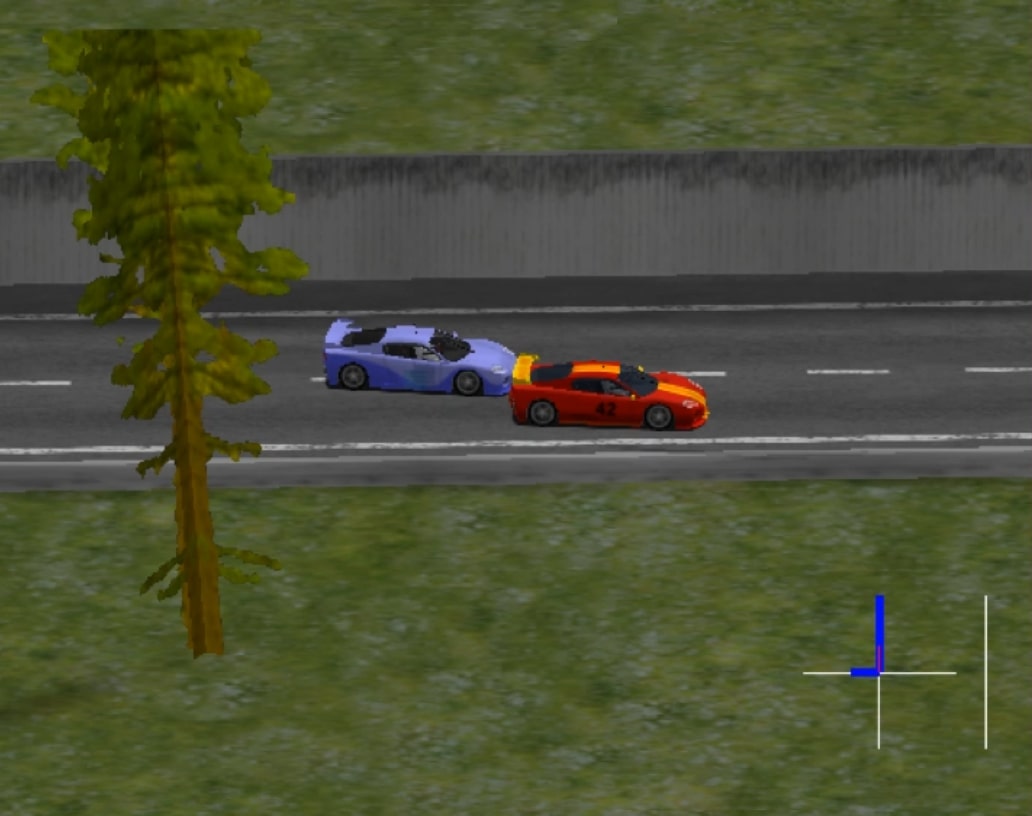}
    \includegraphics[width=0.24\textwidth,height=8.5em]{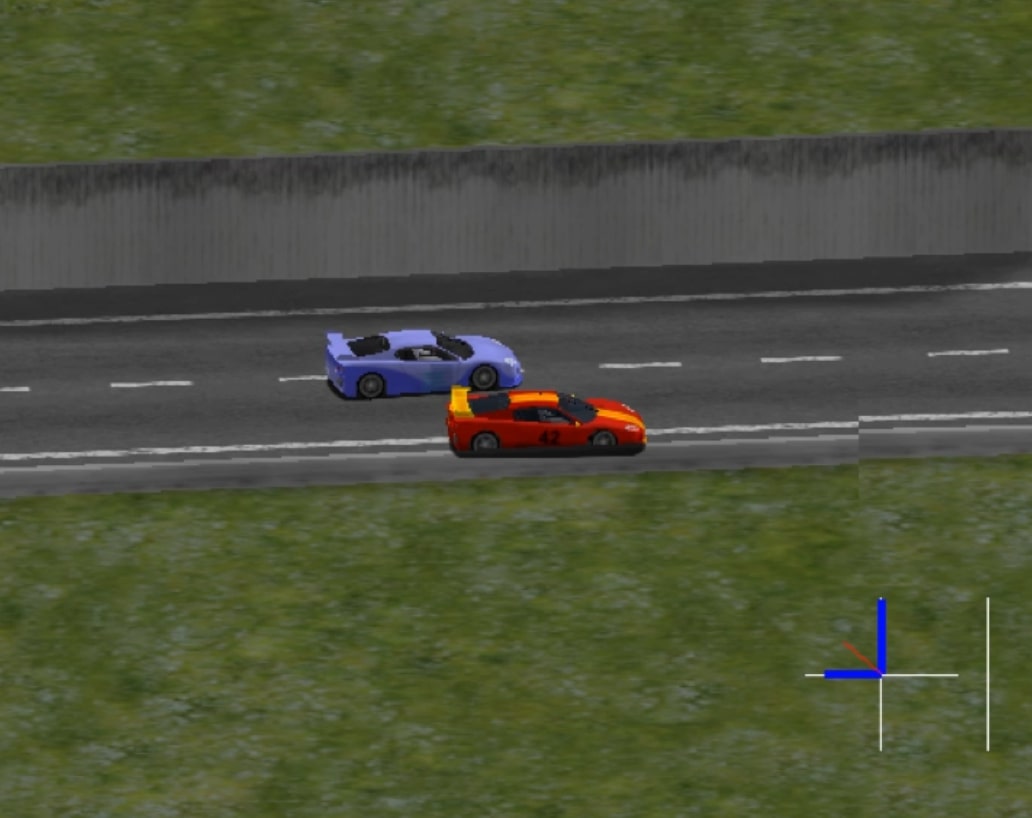}
    \includegraphics[width=0.24\textwidth,height=8.5em]{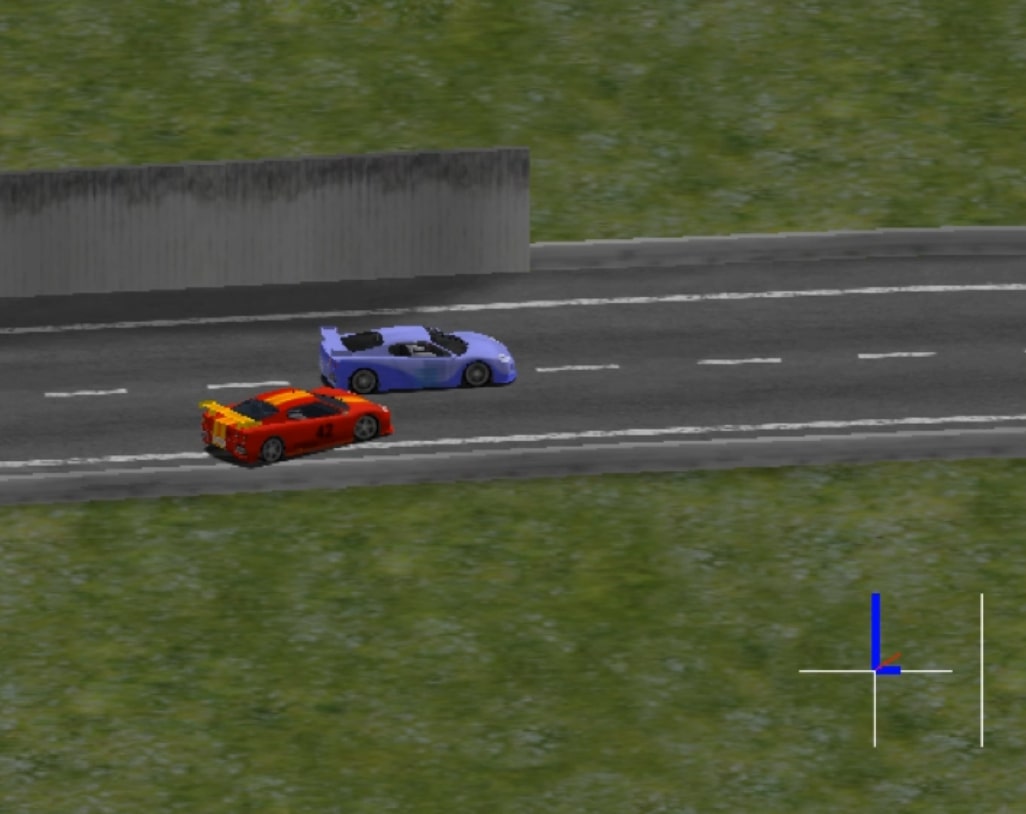}
    \includegraphics[width=0.24\textwidth,height=8.5em]{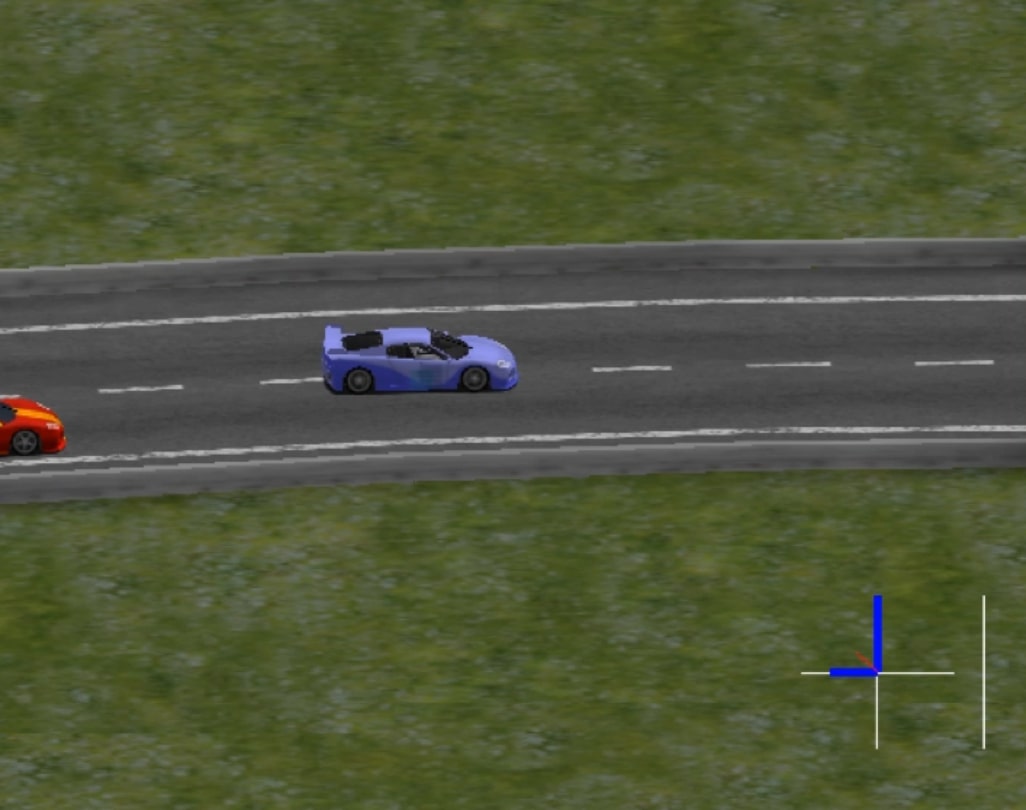}
  \caption{Compete Mode: our car (blue) over take competitor (orange) after a S-curve. For a complete video, please visit \url{https://www.dropbox.com/s/balm1vlajjf50p6/drive4.mov?dl=0}.}
  \label{fig:scurve}
\end{figure}
We illustrate 12 consecutive images in Figure \ref{fig:scurve} to show how our agent (blue) round a s-curve and how the overtake happens. Notably, TORCS has embedded a good physics engine and models vehicle drifting when the speed is fast. We found that the drifting is the main reason of driving in wrong direction after passing a corner and causes terminating the episode early. To deal with this issue, our agent has to decrease the speed before turning, either by hitting the brake or releasing the accelerator, which is also how people drive in real life. After training, we found our model do learned to release the accelerator to slow down before the corner to avoid drifting. Also, from Figure \ref{fig:scurve} we can find that our model did not learn how to avoid collision with competitors. This is because in training mode, there is no competitors introduced to the environment. Therefore, even our car (blue) can passing the s-curve much faster than the competitor (orange), without actively making a side-over-take, our car got blocked by the orange competitor during the s-curve, and finished the overtaking after the s-curve. We witnessed lots of overtakes near or after turning points, this indicates our model works better in dealing with curves. Usually after one to two circles, our car took the first place among all competitors. We uploaded the complete video at \href{https://www.dropbox.com/s/balm1vlajjf50p6/drive4.mov?dl=0}{Dropbox}.

\begin{figure}[!t]
  \centering
  \includegraphics[width=\textwidth,height=0.6\textheight]{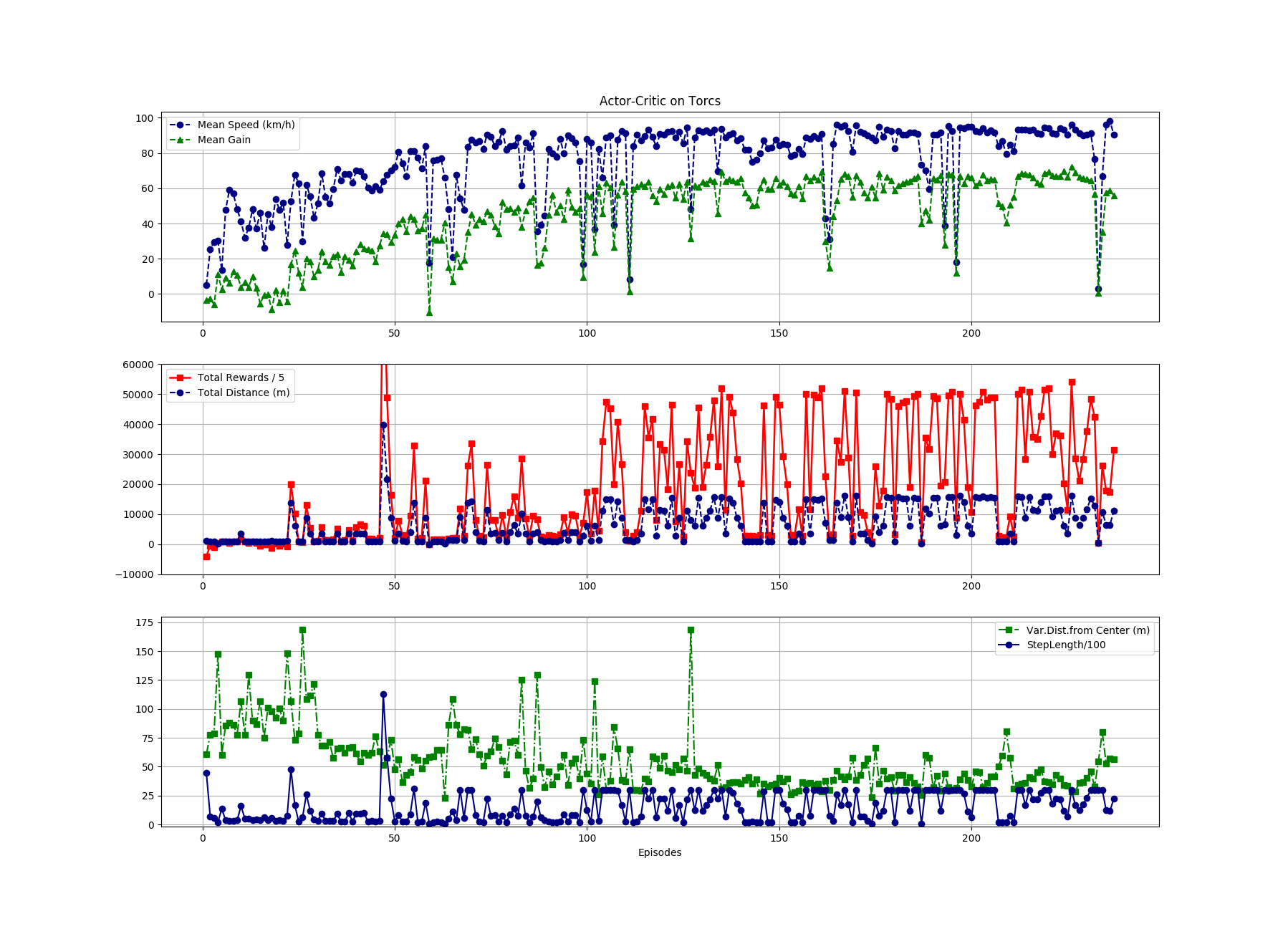}
  \vspace{-1.5cm}
  \caption{Model performance in episodes}
  \label{fig:model_perform}
\end{figure}

We plot the performance of the model during the training in Figure \ref{fig:model_perform}, which contains 3 sub-figures and we refer them from top to bottom as (top), (mid), (bottom). The x-axis of all 3 sub-figures are aligned episodes of training.

In Figure \ref{fig:model_perform}(top), the mean speed of the car (km/h) and mean gain for each step of each episodes were plotted. Specifically, speed of the car is only calculated the speed component along the front direction of the car. In other words, drifting speed is not counted. The gain for each step is calculated with eq.(10). From the figure, as training went on, the average speed and step-gain increased slowly, and stabled after about 100 episodes. This indicates the training actually get stabled after about 100 episodes of training. Apart from that, we also witnessed simultaneously drop of average speed and step-gain. This is because even after the training is stale, the car sometimes could also rushed out of track and got stuck. Normally, when the car rushed out of the track, the episode should terminate. However, it did not guarantee successful termination every time, and this might because of imprecise detection of this out-of-track in TORCS. When the stuck happens, the car have 0 speed till and stuck up to 60000 iterations, and severely decreased the average speed and step-wise gain of this episode. Also, lots of junk history from this episode flush the replay buffer and unstabilized the training. Since this problem originates in the environment instead of in the learning algorithm, we did not spent too much time to fix it, but rather terminated the episode and continue to next one manually if we saw it happen.

In Figure \ref{fig:model_perform}(mid), we plot the total travel distance of our car and total rewards in current episode, against the index of episodes. Intuitively, we can see that as training continues, the total reward and total travel distance in one episode is increasing. This is because the model was getting better, and less likely crash or run out track. Ideally, if the model is optimal, the car should run infinitely, and the total distance and total reward would be stable. However, because of the same reason we mention above, we constantly witness the sudden drop. Notably, most of the "drop" in "total distance" are to the same value, this proves for many cases, the "stuck" happened at the same location in the map.

In Figure \ref{fig:model_perform}(bottom), we plot the variance of distance to center of track (Var.Dist.from.Center(m)), and step length of one episode. The variance of distance to center of the track measures how stable the driving is. We show that our trained agent often drives like a "drunk" driver, in 8-shape, at the beginning, and gradually drives better in the later phases. The Var.Dist.from.Center curve decreases and stabilizes after about 150 episodes. This indicates the our model still drive unstably after 100 episodes, when the speed and episode rewards already get stabilized. So the extra 50 episodes of training stabilize the training.

\section{Conclusion}
In order to bridge the gap between autonomous driving and reinforcement learning, we adopt the deep deterministic policy gradient (DDPG) algorithm to train our agent in The Open Racing Car Simulator (TORCS). In particular, we select appropriate sensor information from TORCS as our inputs and define our action spaces in continuous domain. We then design our rewarder and network architecture for both actor and critic inside DDPG paradigm. We demonstrate that our agent is able to run fast in the simulator and ensure functional safety in the meantime. 

{\small
\bibliographystyle{ieee}
\bibliography{egbib}
}
\end{document}